\documentclass[10pt,twocolumn,letterpaper]{article}

\usepackage{cvpr}              
\definecolor{cvprblue}{rgb}{0.21,0.49,0.74}
\usepackage[pagebackref,breaklinks,colorlinks,allcolors=cvprblue]{hyperref}
\usepackage{url}
\usepackage{arydshln}
\usepackage{algorithm}
\usepackage{algorithmic}
\usepackage{stfloats}
\usepackage{xcolor}
\usepackage[table]{xcolor}
\usepackage{svg}
\usepackage{wrapfig}
\usepackage{graphicx}
\usepackage{amsmath}
\usepackage{amssymb}
\usepackage{booktabs}
\usepackage[dvipsnames]{xcolor}
\usepackage{booktabs}
\usepackage{colortbl}
\usepackage{enumitem}

\usepackage[capitalize]{cleveref}
\crefname{figure}{Figure}{Figures}
\Crefname{section}{Section}{Sections}
\Crefname{table}{Table}{Tables}
\newlength\savewidth
\newcolumntype{L}[1]{>{\raggedright\arraybackslash}p{#1}}  
  
\usepackage[table,xcdraw]{xcolor}
\newcommand{\ourmethod}{\textit{RAS}}

\def\paperID{} 
\def\confName{CVPR}
\def\confYear{2026}

\title{Region-Adaptive Sampling for Diffusion Transformers}

\author{Ziming Liu\thanks{Equal contribution}\\
National University of Singapore\\
{\tt\small e0732706@u.nus.edu}
\and
Yifan Yang\footnotemark[1]\\
Microsoft Research\\
{\tt\small yifanyang@microsoft.com}
\and
Chengruidong Zhang\\
Microsoft Research\\
{\tt\small chengzhang@microsoft.com}
\and
Yiqi Zhang\\
National University of Singapore\\
{\tt\small yiqi.zhang@comp.nus.edu.sg}
\and
Lili Qiu\\
Microsoft Research\\
{\tt\small liliqiu@microsoft.com}
\and
Yang You\thanks{Corresponding Authors}\\
National University of Singapore\\
{\tt\small youy@comp.nus.edu.sg}
\and
Yuqing Yang\footnotemark[2]\\
Microsoft Research\\
{\tt\small yuqyang@microsoft.com}
}

\begin{document}
\maketitle
\begin{abstract}
Diffusion models (DMs) have achieved state-of-the-art performance across diverse generative tasks, yet their dependence on sequential forward passes fundamentally limits real-time efficiency. Existing acceleration methods primarily focus on reducing the number of sampling steps or reusing intermediate features. Leveraging the inherent flexibility of Diffusion Transformers (DiTs) in handling variable token counts, we introduce \textbf{\ourmethod{}}, a training-free sampling strategy that dynamically adjusts update ratios across image regions based on model attention. Our key insight is that DiTs progressively focus on semantically meaningful regions, and such focused areas exhibit strong temporal continuity between consecutive steps. Building on this observation, \ourmethod{} updates only these focused regions while reusing cached noise elsewhere, with focus maps inferred from the previous step’s output. Experiments on \textit{Stable Diffusion 3} and \textit{Lumina-Next-T2I} demonstrate up to \textbf{2.36$\times$} and \textbf{2.51$\times$} speedups, respectively, with negligible quality degradation—highlighting a practical pathway toward real-time diffusion transformer generation.

\end{abstract}    
\section{Introduction}
\label{sec:intro}
Diffusion models (DMs)~\cite{Ho2020diffusion, Prafulla2024diffusionbeats, Yang2019Generative, Jascha2015deep} have emerged as highly effective probabilistic generative models, producing high-quality data across a wide range of domains. They have been successfully applied to image synthesis~\cite{rombach2022high, dhariwal2021diffusion}, image super-resolution~\cite{li2022srdiff, yue2024resshift, gao2023implicit}, image-to-image translation~\cite{wang2022pretraining, saharia2022palette, li2023bbdm}, image editing~\cite{kawar2023imagic, zhang2023sine}, inpainting~\cite{lugmayr2022repaint}, video synthesis~\cite{blattmann2023align, esser2023structure}, text-to-3D generation~\cite{poole2022dreamfusion}, and even planning tasks~\cite{janner2022planning}.  
Despite their impressive generative capabilities, sampling from DMs requires solving a stochastic or ordinary differential equation (SDE/ODE)~\cite{protter2005stochastic, hartman2002ordinary} in reverse time, which entails multiple sequential forward passes through a large neural network. This inherent sequential dependency significantly limits their real-time applicability.

\begin{figure}
    \centering
        \includegraphics[width=\linewidth]{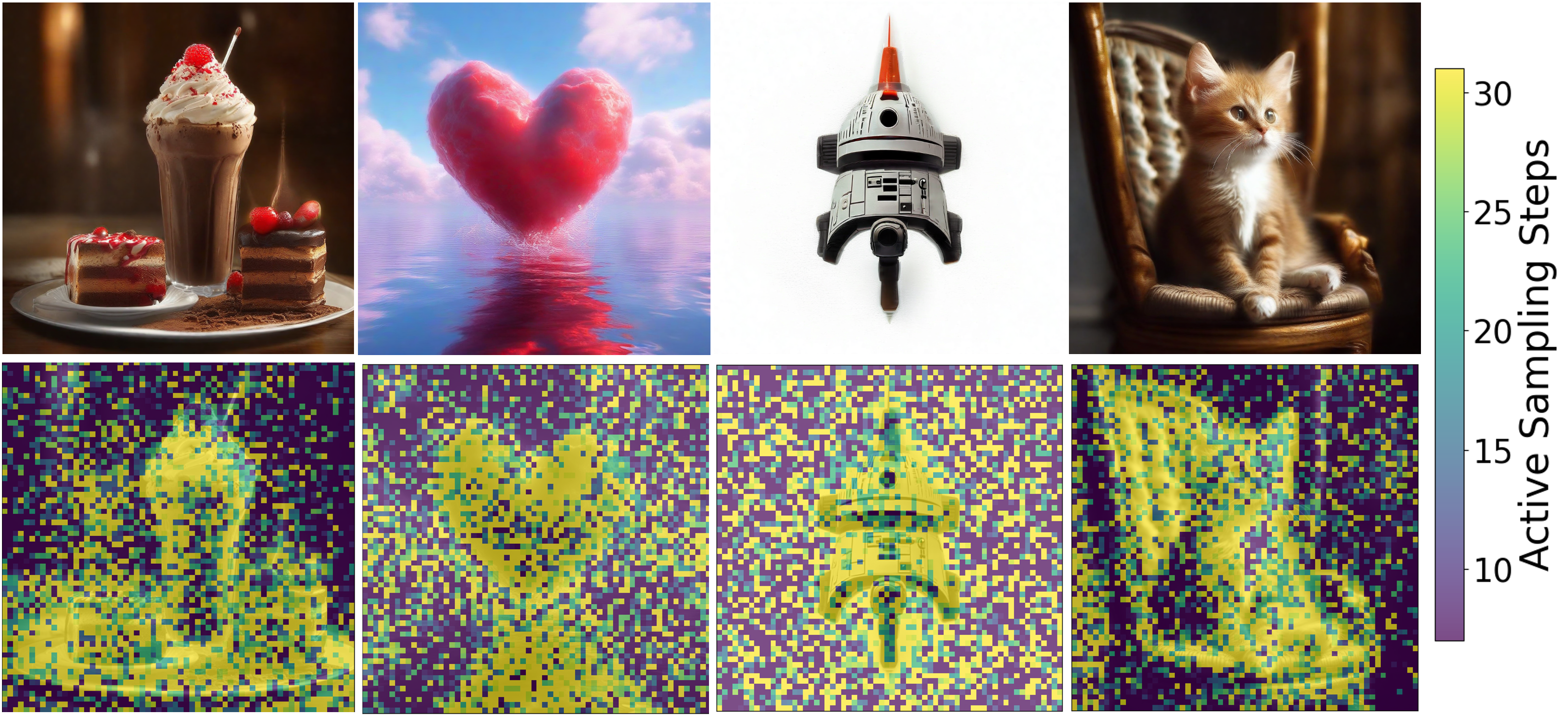}
        \caption{The main subject and the regions with more details are \textit{brushed} for more steps than other regions in \ourmethod{}. Each block represents a patchified latent token.}
        \label{fig:drop_cnt_map}
\end{figure}

Considerable effort has been devoted to accelerating the sampling process in diffusion models (DMs) by reducing the number of sampling steps. Existing approaches can be broadly categorized into training-based methods, such as progressive distillation~\cite{salimansprogressive}, consistency models~\cite{song2023consistency}, and rectified flow~\cite{liu2022flow, albergo2022building, lipman2022flow}, and training-free methods, including DPM-Solver~\cite{lu2022dpm}, AYS~\cite{sabouralign}, DeepCache~\cite{xu2018deepcache}, and Delta-DiT~\cite{chen2024delta}. However, these techniques process all regions of an image uniformly during sampling, regardless of regional content complexity or semantic importance.  

\begin{figure*}[h]
    \centering
    \includegraphics[width=0.8\textwidth]{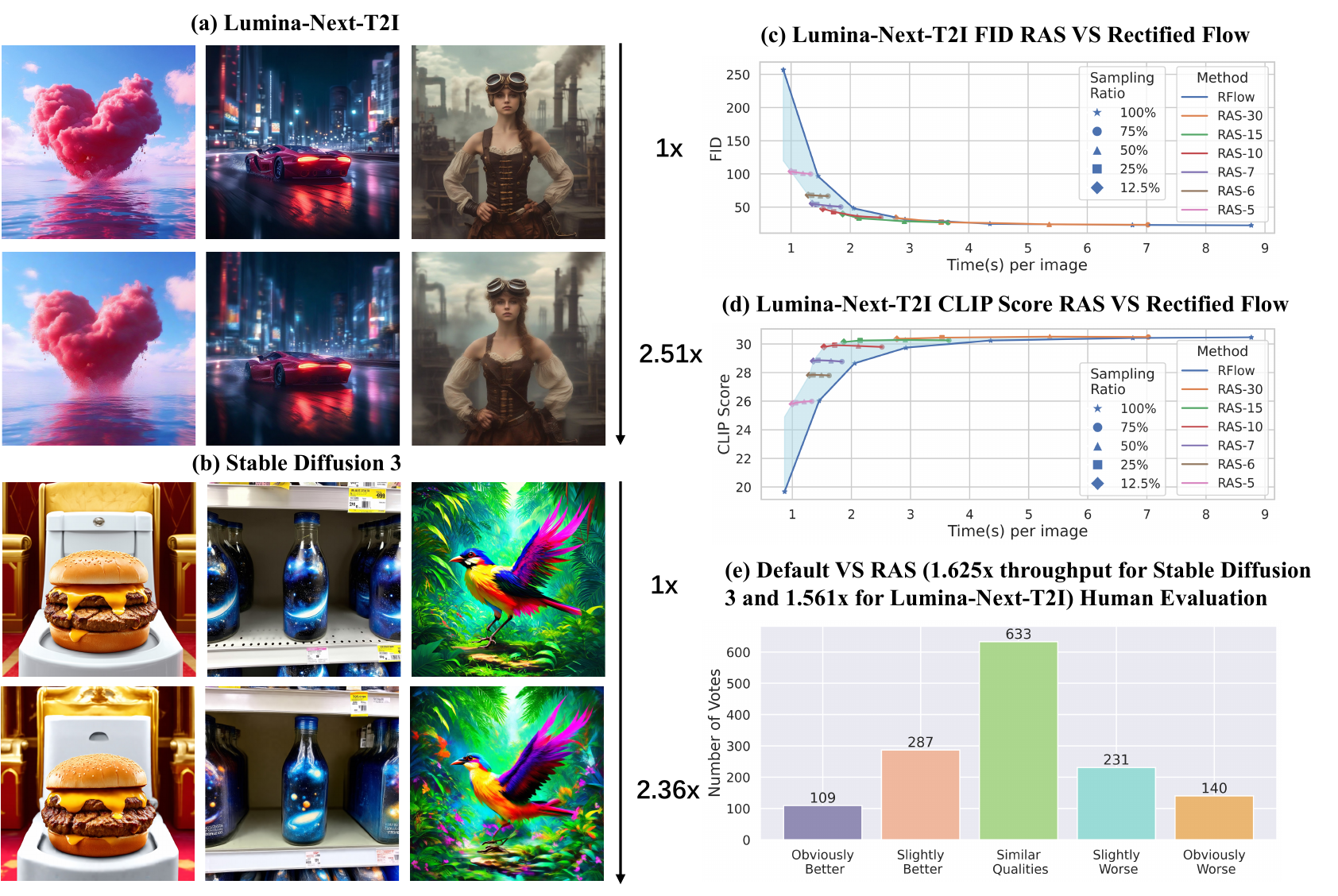}

    \caption{(a)(b) Acceleration results on Lumina-Next-T2I and Stable Diffusion~3 using 30 and 28 timesteps, respectively. 
    (c)(d) Multiple configurations of RAS outperform rectified flow in both visual fidelity and text alignment; “RAS-X’’ denotes RAS with $X$ total sampling steps. 
    (e) RAS achieves comparable human evaluation scores to the default models while providing $\sim1.6\times$ higher throughput.}
    \label{fig:teaser}
\end{figure*}
Intuitively, different regions within an image exhibit varying levels of structural and semantic complexity: intricate foreground details may require more refinement steps to preserve fidelity, while smooth or repetitive background areas could tolerate more aggressive step reduction without noticeable quality loss. This observation motivates the development of a more flexible sampling strategy that dynamically adjusts the update ratio across regions, achieving faster generation while maintaining perceptual quality.

This direction represents a natural progression in the evolution of diffusion models. From DDPM~\cite{Ho2020diffusion} to Stable Diffusion XL~\cite{podell2023sdxlimprovinglatentdiffusion}, most existing models have been built upon U-Net backbones~\cite{ronneberger2015u}, whose convolutional structures impose uniform spatial processing due to their fixed-size grid inputs. In contrast, the emergence of Diffusion Transformers (DiTs)~\cite{William2023DiT} and the growing adoption of fully transformer-based generative architectures~\cite{vaswani2017attention} enable flexible tokenization and non-uniform computation across spatial regions. Leveraging this property, we are inspired to design a novel sampling strategy that adaptively allocates different sampling steps to different image regions, paving the way for more efficient and semantically aware diffusion generation.

\begin{figure*}[t]
    \centering
    \includegraphics[width=0.9\linewidth]{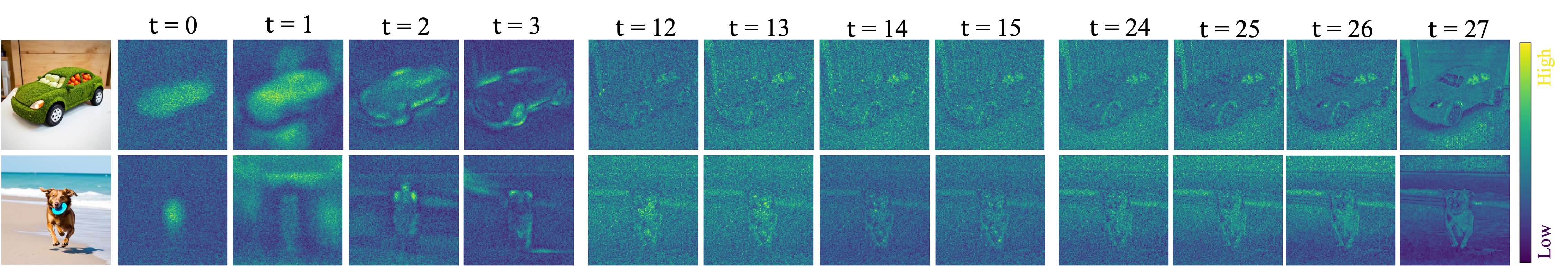}
    \caption{
        Visualization of predicted noise across diffusion steps. 
        The DiT model gradually focuses on semantically meaningful regions at each timestep, 
        and the attention shift between steps exhibits strong temporal continuity.
    }
    \label{fig:latents}
\end{figure*}

To evaluate the feasibility of this idea, we visualized the intermediate diffusion outputs at various sampling steps (Figure~\ref{fig:latents}). Two clear patterns emerged: (1) regions of focus exhibit strong continuity across adjacent steps in later stages, and (2) at each step, the model predominantly concentrates on semantically meaningful areas of the image. This behavior resembles an artist refining a canvas through thousands of careful strokes, where each iteration selectively enhances specific regions. Consequently, regions that receive little attention at a given step could potentially be skipped during computation in Diffusion Transformers (DiTs), enabling the model to concentrate computational resources on regions of interest.

\begin{figure}[t]
    \centering
    \includegraphics[width=0.8\linewidth]{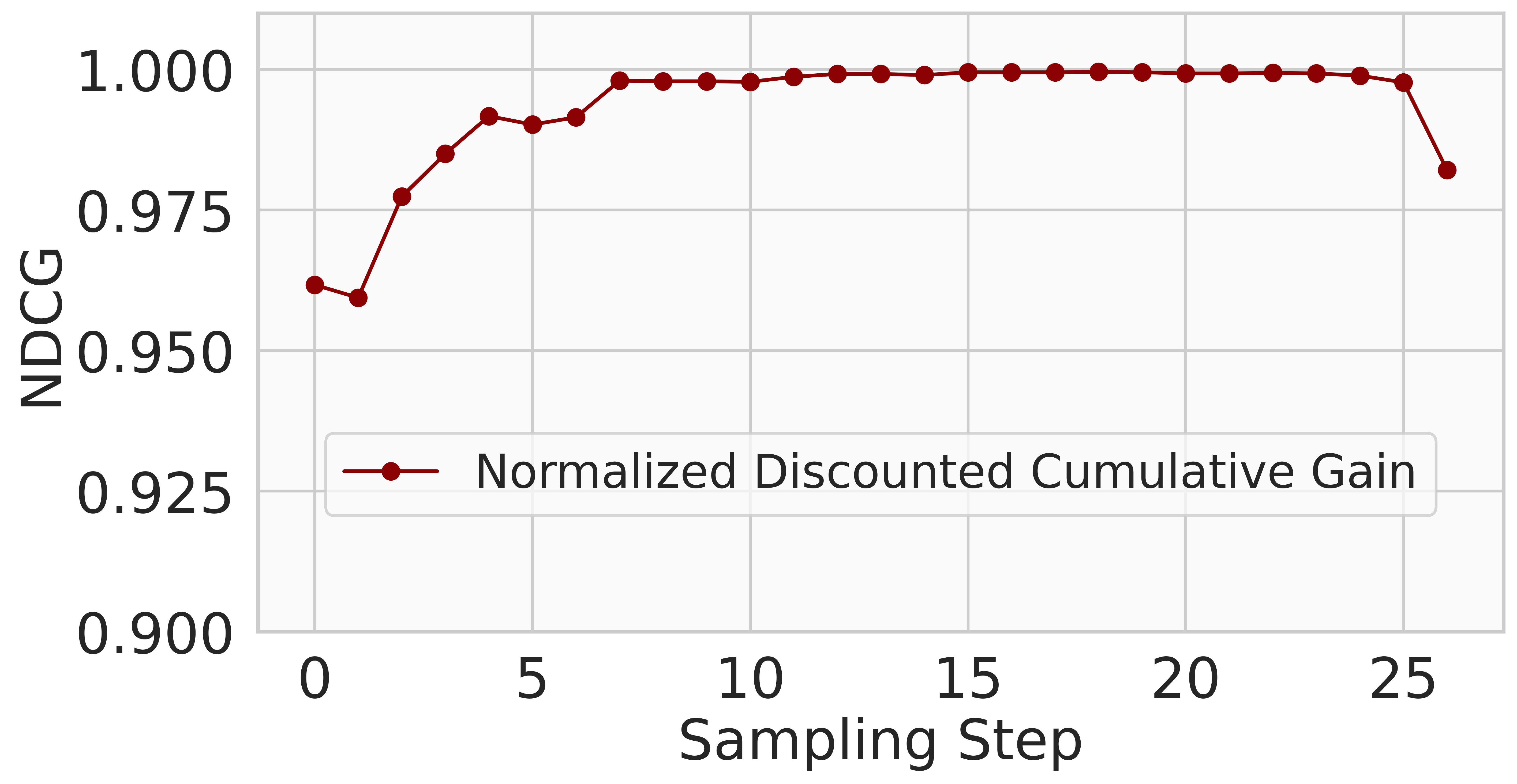}
    \caption{
        Normalized Discounted Cumulative Gain (NDCG)~\cite{dcg, wang2013theoreticalanalysisndcgtype} 
        between adjacent sampling steps. 
        The consistently high NDCG values throughout the diffusion process 
        indicate strong similarity in the ranking of focused tokens 
        (ranging from 0 to 1), reflecting temporal continuity of model attention.
    }
    \label{fig:ndcg}
\end{figure}

We further validated this hypothesis by ranking tokens at each step using our proposed \emph{output-noise metric}, which identifies the areas that the model focuses on most. We then measured the ranking similarity between consecutive steps using the NDCG metric (Figure~\ref{fig:ndcg}), revealing high temporal consistency in attention across steps. These findings motivate a sampling strategy that adaptively allocates different update ratios to regions based on their attention persistence, paving the way for efficient yet high-quality diffusion sampling.

\begin{figure}[t]
    \centering
    \includegraphics[width=0.65\linewidth]{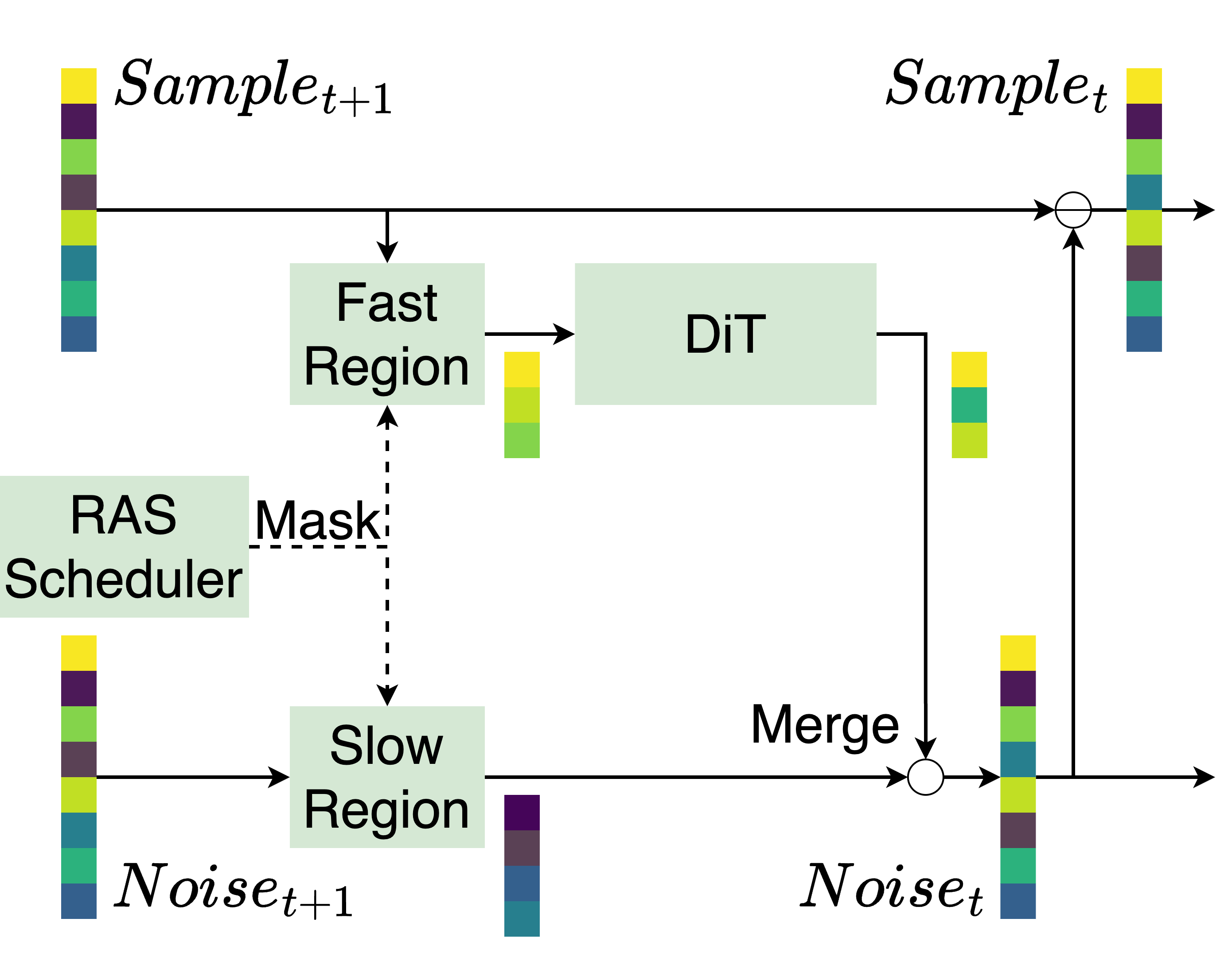}
    \caption{
        Overview of the \textbf{\ourmethod{}} framework. 
        At each diffusion step, only the \emph{fast-update} regions identified from the previous output 
        are forwarded through the model, while the remaining regions reuse cached noise. 
        This selective update mechanism enables region-adaptive computation and efficient denoising.
    }
    \label{fig:drop_overview}
\end{figure}

As illustrated in Figure~\ref{fig:drop_overview}, our method leverages the output noise from the previous step to identify the model’s primary focus for the current step, referred to as the \emph{fast-update regions}. Only these regions are forwarded through the Diffusion Transformer (DiT) for denoising, while the remaining \emph{slow-update regions} reuse the cached noise from the previous step. This design introduces regional variability in the number of effective sampling steps: regions of interest are updated more frequently, whereas less critical areas retain their previous noise estimates, thereby reducing overall computation.  

For each input \( X_t \), a fast-update rate determines the proportion of tokens selected for refinement. The updated fast-region noise is then combined with the preserved slow-region noise to construct the next-step input \( X_{t-1} \). To maintain global semantic and visual consistency, features from slow-update regions are retained as reference keys and values for subsequent attention computations. Although fast-region selection is dynamic and recalculated at every step to prioritize semantically significant areas, we periodically perform full-image updates to prevent cumulative approximation errors.  
In summary, we propose \textbf{\ourmethod{}}, the first diffusion sampling strategy that enables \emph{regionally adaptive sampling ratios}. Unlike spatially uniform samplers, our approach allocates DiT’s computation to regions of current importance, achieving a better balance between efficiency and fidelity. As shown in Figure~\ref{fig:teaser}(c)(d), \ourmethod{} substantially reduces inference cost with minimal FID degradation, while surpassing uniform baselines in FVD under equivalent budgets. Furthermore, Figure~\ref{fig:teaser}(a)(b) demonstrates that with large-scale models such as Lumina-Next-T2I~\cite{gao2024luminat2xtransformingtextmodality} and Stable Diffusion~3~\cite{esser2024sd3}, our adaptive fast-region updating yields more than \textbf{2$\times$} acceleration with negligible perceptual quality loss.

\section{Related Work}
\label{sec.back}

\subsection{Diffusion Models: From U-Net to Transformer}

Diffusion models (DMs)~\cite{Ho2020diffusion, Prafulla2024diffusionbeats, Yang2019Generative, Jascha2015deep} have demonstrated remarkable generative capabilities, often surpassing generative adversarial networks (GANs)~\cite{goodfellow2014generative} across diverse downstream tasks. Early variants such as DDPMs~\cite{Ho2020diffusion} and Stable Diffusion XL~\cite{podell2023sdxlimprovinglatentdiffusion} primarily adopted convolutional U-Net backbones~\cite{ronneberger2015u}. However, convolutional architectures inherently preserve strict spatial locality and resolution for operations like pooling, which constrains their ability to exploit redundancy in latent representations and complicates pruning or selective computation.

This limitation has been addressed by the advent of Diffusion Transformers (DiTs)~\cite{William2023DiT}, now employed in state-of-the-art generative models including Stable Diffusion~3~\cite{esser2024scalingrectifiedflowtransformers}, Lumina-T2X~\cite{gao2024luminat2xtransformingtextmodality}, and PixArt-$\Sigma$~\cite{chen2024pixartsigmaweaktostrongtrainingdiffusion}. In contrast to U-Nets, DiTs adopt a fully Transformer-based architecture~\cite{vaswani2017attention} enhanced with adaptive layer normalization for conditional prompts, thereby eliminating convolution entirely. Positional information is encoded via embeddings, making latent tokens spatially independent and enabling flexible token-level manipulation. This property allows us to exploit redundancy (Section~\ref{sec:intro}) by selectively updating the most relevant tokens at each sampling step while reusing cached noise predictions for others.

\subsection{Efficient Diffusion Model Inference}
Reducing the high computational cost of diffusion model inference has been an active area of research. A common line of work focuses on decreasing the number of sampling steps required for generation. Several methods achieve this through additional training, such as progressive distillation~\cite{salimansprogressive}, consistency models~\cite{song2023consistency}, and rectified flow~\cite{liu2022flow, lipman2022flow, albergo2022building}.  
Progressive distillation iteratively applies a specialized distillation process that transfers a pretrained deterministic sampler into one capable of using substantially fewer diffusion steps. Latent Consistency Models (LCMs), in contrast, directly predict the clean image instead of the noise component during inference, allowing high-quality image synthesis with drastically reduced iterations.  
Among these, rectified flow has been particularly influential and is adopted in models such as Stable Diffusion~3~\cite{esser2024scalingrectifiedflowtransformers}. It learns an ordinary differential equation (ODE) trajectory that follows a straight path between the standard normal prior and the target data distribution. Such rectified trajectories effectively shorten the transport distance between distributions, thereby lowering the number of required sampling steps and improving inference efficiency.

Training-free methods have also been introduced to reduce either the number of sampling steps or the computational cost per step. For instance, DeepCache~\cite{xu2018deepcache}, designed for U-Net-based diffusion models, caches and reuses intermediate features across adjacent stages to skip redundant down- and up-sampling operations. However, such approaches uniformly process all image regions, disregarding the heterogeneous complexity present across different spatial areas, which leads to suboptimal efficiency.

As discussed in Section~\ref{sec:intro}, image regions often vary significantly in structural and semantic complexity. To exploit this heterogeneity, we propose \textbf{\ourmethod{}}, which adapts the computational allocation dynamically based on region-specific characteristics. Unlike existing methods that uniformly update all regions, \ourmethod{} selectively refines the model’s regions of focus while reusing cached noise for less salient areas. Notably, \ourmethod{} is orthogonal to prior acceleration techniques, including step-reduction methods and module-level optimizations such as DiTFastAttn~\cite{yuan2024ditfastattnattentioncompressiondiffusion} and $\Delta$-DiT~\cite{chen2024delta}, and can be seamlessly combined with them for further efficiency gains.

\section{Methodology}
\label{sec.method}

\begin{table}

\centering
\renewcommand\arraystretch{1.2}
\label{tab.symbo}
\scalebox{0.85}{\begin{tabular}{ll}
\noalign{\hrule height 1pt}
$t$    & The current timestep                                            \\
$N$    & The noise output of the DiT model              \\
$\widetilde{N}$    & The cached noise output from the previous timestep                                       \\
$\hat{N}$    & The estimated full-length noise calculated with $N$ and $\widetilde{N}$\\
$S$ & The unpathified image sample \\
$x$ & The pathified input of the DiT model \\
$M$ & Mask generated to drop certain tokens in the input \\
$D$ & The number of times the tokens in a patch being dropped \\
\noalign{\hrule height 1pt}
\end{tabular}
}
\caption{Notation summary.}
\end{table}

\subsection{Overview}
In this section, we introduce the overall design of \textbf{\ourmethod{}} and describe the techniques used to exploit inter-timestep token correlations and the regional token attention mechanism introduced in Section~\ref{sec:intro}. 
Our framework is built upon three key components:  
(1) Motivated by the regional characteristics observed during DiT inference, we design an end-to-end pipeline that dynamically eliminates DiT computation for selected tokens at each timestep;  
(2) To capture temporal continuity across consecutive timesteps, we propose a simple yet effective method for identifying \emph{fast-update regions} that require refinement in subsequent steps; and  
(3) Based on our observations of consistent spatial distribution patterns, we introduce several scheduling optimization techniques to further enhance the fidelity of the generated results.

\subsection{Region-Adaptive Sampling}

\textbf{Region-Aware DiT Inference with \ourmethod{}.}  
Building on the insight that only certain regions are critical at each timestep, we design the \ourmethod{} inference pipeline for Diffusion Transformers (DiTs).  
In U-Net–based diffusion models such as SDXL~\cite{podell2023sdxlimprovinglatentdiffusion}, token positions must remain fixed to preserve spatial structure.  
In contrast, the architecture of DiT, where positional information is injected via embeddings such as RoPE~\cite{su2024roformer}, allows masking or reordering of latent tokens without disrupting positional encoding.  
This flexibility enables us to selectively determine which regions are actively processed by the model.

At the end of each timestep, the current latent sample is reconstructed by merging the newly generated outputs for active tokens with the cached noise from inactive tokens.  
Formally, the noise sequence is restored by integrating the model output for the fast-update regions with the previously cached noise for the slow-update regions.  
This mechanism allows important tokens to move toward the updated direction determined at the current timestep, while less critical tokens maintain their previous trajectories.

To facilitate this process, we compute a region-wise metric \( R \) to identify fast-update regions based on the model’s output noise.  
We then update the drop count \( D \) to record how frequently each token has been skipped, and generate a binary mask \( M \) that governs computation in the subsequent step.  
Using \( M \), the noise of slow-update regions is cached, while the tokens in fast-update regions are patchified and passed through the DiT model.  
Since modules such as LayerNorm~\cite{ba2016layernormalization} and MLP operate independently across tokens, their computation remains consistent even when the token sequence is incomplete.  
For the attention module~\cite{vaswani2017attention}, we further introduce a caching mechanism to accelerate repeated key–value lookups, as detailed in a later section.  
Overall, \ourmethod{} dynamically detects regions of focus and reduces DiT’s computational workload by at least the same proportion as the user-defined sampling ratio.

\noindent \textbf{Region Identification.}  
At each timestep, the DiT model receives the current timestep embedding, latent sample, and prompt embedding to predict the noise guiding the sample toward the clean image.  
To quantify the refinement requirement of each token, we analyze the model’s noise output and observe that the \emph{standard deviation} of predicted noise effectively distinguishes different semantic regions.  
Empirically, the main subject (fast-update regions) exhibits noticeably lower noise variance than the background (slow-update regions), likely reflecting the uneven information density after the addition of Gaussian noise.  
Using the standard deviation as a region-selection metric yields robust results, highlighting semantically meaningful regions, maintaining image quality, and producing distinct contrasts between focused and background areas.

\noindent \textbf{Temporal Token Continuity.}  
Given the strong similarity between latent samples across adjacent timesteps, we hypothesize that tokens deemed important at the current timestep are likely to remain important in subsequent ones. Conversely, less-focused tokens can often be safely dropped with minimal perceptual impact.  
Before presenting the final formulation of our importance metric, we first introduce a mechanism to prevent the persistent exclusion of certain regions.

\noindent \textbf{Starvation Prevention.}  
During the diffusion process, the primary subject regions typically require more frequent refinement than background areas.  
However, repeatedly skipping background tokens may lead to excessive blurring or noise accumulation in the final output.  
To mitigate this issue, we track the frequency with which each token is dropped and incorporate this \emph{drop count} as a scaling factor within our region-selection metric.  
This adjustment ensures that even low-importance tokens are periodically revisited, preventing starvation and maintaining global consistency throughout sampling.

Since the Diffusion Transformer (DiT) processes patchified latent tokens, we compute our metric at the patch level by averaging the per-token scores within each patch.  
Combining the above factors, our final region-selection metric is defined as:
\begin{equation}
\label{equa:metric}
    R_t = \mathrm{mean}_{\text{patch}} \big( \mathrm{std}(\hat{N}_t) \big) \cdot \exp(k \cdot D_{\text{patch}}),
\end{equation}
where $\hat{N}_t$ denotes the predicted noise at timestep $t$, $D_{\text{patch}}$ is the number of times the tokens in a patch have been dropped, and $k$ is a scaling coefficient that controls the contrast between fast-update and slow-update regions.  
A higher $k$ value enforces more aggressive recovery of long-inactive patches, balancing efficiency with image quality.

\noindent \textbf{Key–Value Caching for Attention.}  
The self-attention mechanism computes relationships between all tokens by using their queries, keys, and values.  
In \ourmethod{}, the attention of active tokens can in principle be computed using only other active tokens.  
However, our metric $R_t$ determines active and inactive regions solely based on noise statistics, without explicitly accounting for their contributions to the attention context.  
Naively excluding inactive tokens would therefore distort attention outputs and degrade generation quality.

To preserve contextual integrity while maintaining efficiency, we introduce a \emph{key–value caching} mechanism.  
During each step, the complete key and value tensors are cached, and only the portions corresponding to active tokens are updated.  
This approach leverages the temporal consistency between adjacent timesteps: since token embeddings evolve smoothly, previously cached representations for inactive regions remain strong approximations of their true values.  
The attention output for the active tokens can thus be estimated as:
\begin{equation}
    O_{a} = \mathrm{softmax}\!\left(\frac{Q_{a}[K_{a}, \widetilde{K}_{i}]^{\top}}{\sqrt{d}}\right)[V_{a}, \widetilde{V}_{i}],
\end{equation}
where $Q_{a}$, $K_{a}$, and $V_{a}$ denote the query, key, and value matrices of the active tokens, while $\widetilde{K}_{i}$ and $\widetilde{V}_{i}$ represent the cached keys and values of the inactive tokens.  
This formulation approximates the full-attention output with minimal computation and negligible quality degradation. Please refer to the appendix for more details.
\subsection{Scheduling Optimization}

\noindent \textbf{Dynamic Sampling Ratio.}  
As shown in Figure~\ref{fig:ndcg}, correlations between adjacent timesteps are relatively weak during the early diffusion stages but gradually strengthen as the process stabilizes, consistent with the patterns observed in Figure~\ref{fig:latents}.  
This suggests that applying selective sampling too early may damage the structural foundation of the generated image.  
To address this, we employ a \emph{dynamic sampling schedule}: the initial few steps (e.g., the first 4 out of 28) are executed with a full 100\% sampling ratio to preserve the global image outline, after which the ratio is gradually reduced as the generation stabilizes.  
This adaptive design balances quality and efficiency, enabling substantial computational savings while minimizing any degradation in fine-grained details.

\noindent \textbf{Accumulated Error Resetting.}  
Because \ourmethod{} focuses on regions of interest that persist across adjacent sampling steps, tokens in less-attended regions may remain inactive for extended periods.  
Without intervention, these regions can accumulate stale denoising directions, leading to noticeable discrepancies between the latent produced by \ourmethod{} and that from the original full sampling process.  
To mitigate this, we introduce \emph{dense steps}, periodic full updates inserted into the diffusion process to reset accumulated errors.  
For example, in a 30-step sampling schedule where \ourmethod{} begins at step~4, we designate steps~12 and~20 as dense steps.  
During these steps, all regions are reprocessed by the model, allowing corrections to any drift that may have developed in inactive areas.  
This periodic reset mechanism ensures long-term stability and keeps the denoising trajectory aligned with the intended generative path.

\subsection{Implementation}

\noindent \textbf{Kernel Fusing.}  
To reduce redundant computation, we employ key--value caching in self-attention and update only the active tokens during selective sampling steps.  
These partial updates correspond to a scatter operation on active-token indices.  
Instead of launching a separate scatter kernel, we fuse this operation into the preceding GeMM kernel, enabling in-place updates and avoiding extra memory copies.  
Following the principle in PIT~\cite{zheng2023pit}, where permutation-invariant transformations can be integrated into GPU I/O stages with negligible overhead, we embed the scatter in the GeMM epilogue, effectively merging linear projection and token reindexing into a single efficient operation.  
This fusion eliminates synchronization and I/O overhead, yielding notable latency reductions during \ourmethod{} inference.

\section{Experiments}
\label{sec.exp}

\subsection{Experiment Setup}

\noindent \textbf{Models, Datasets, Metrics, and Baselines.}  
We evaluate \textbf{\ourmethod{}} on two state-of-the-art text-to-image diffusion models: \textit{Stable Diffusion~3}~\cite{esser2024scalingrectifiedflowtransformers} and \textit{Lumina-Next-T2I}~\cite{gao2024luminat2xtransformingtextmodality}.  
Experiments are conducted using 10,000 randomly sampled caption–image pairs from the MS-COCO~2017 dataset~\cite{lin2014microsoftcoco}.  
 To evaluate both visual fidelity and text–image alignment, we adopt three standard metrics:  
(1) \textbf{Fréchet Inception Distance (FID)}~\cite{heusel2017gansfid}, which measures overall image realism;  
(2) \textbf{Sliding FID (sFID)}~\cite{heusel2017gansfid}, which provides a localized perceptual quality assessment; and  
(3) \textbf{CLIP Score}~\cite{hessel2021clipscore}, which quantifies semantic consistency between generated images and their corresponding text prompts. For human evaluation, we have also included ImageReward~\cite{xu2023imagerewardlearningevaluatinghuman}, PickScore~\cite{kirstain2023pickapicopendatasetuser}, and hpsv2 ~\cite{wu2023humanpreferencescorev2} as benchmarks.
 For baseline comparison, we evaluate against a suite of widely adopted rectified-flow and flow-matching–based acceleration methods~\cite{liu2022flow, albergo2022building, esser2024scalingrectifiedflowtransformers, lipman2022flow, dao2023flowmatchinglatentspace, fischer2023boosting}, all of which uniformly reduce the number of diffusion timesteps across the entire image.  
We implement \ourmethod{} under multiple configurations with varying total timestep counts to assess trade-offs between efficiency and image quality, and compare the results against the original full-sampling implementations under equivalent throughput settings.

\noindent \textbf{Code Implementation.}  
We implement \textbf{\ourmethod{}} in \texttt{PyTorch}~\cite{Adam2019PyTorch}, leveraging the \texttt{diffusers} library~\cite{von-platen-etal-2022-diffusers} and its \texttt{FlowMatchEulerDiscreteScheduler} for inference scheduling.  
Evaluation metrics are computed using publicly available implementations from GitHub repositories, including FID~\cite{Seitzer2020FID}, sFID~\cite{Hu2022sFID}, and CLIP Score~\cite{taited2023CLIPScore}.  
All experiments are conducted on four servers, each equipped with eight NVIDIA A100~(40\,GB) GPUs, while latency and throughput benchmarks are measured on a single NVIDIA A100~(80\,GB) GPU.

\begin{table}
\vspace{0.2cm}
\centering
\setlength{\tabcolsep}{1mm}
\fontsize{9pt}{10pt}\selectfont
\setlength\tabcolsep{2pt}
\begin{tabular}{lcccccc}
\hline
\textbf{Method} & \textbf{Steps} & \textbf{Sample} & \textbf{Image/s$\uparrow$} & \textbf{FID $\downarrow$} & \textbf{sFID $\downarrow$} & \textbf{CLIP$\uparrow$} \\
& & \textbf{Ratio} & & & & \textbf{score} \\
\hline
\textbf{SD3} & & & & & & \\
\hline
\rowcolor{gray!10}
RFlow & 5 & 100\% & 1.43 & 39.70 & 22.34 & 29.84 \\
RAS & 7 & 25.0\% & 1.45 & \textbf{31.99} & 21.70 & \textbf{30.64} \\
RAS & 7 & 12.5\% & 1.48 & 32.86 & 22.10 & 30.55 \\
RAS & 6 & 25.0\% & 1.52 & 33.24 & \textbf{21.51} & 30.38 \\
RAS & 6 & 12.5\% & \textbf{1.57} & 33.81 & 21.62 & 30.33 \\
\hline

\rowcolor{gray!10}
RFlow & 4 & 100\% & 1.79 & 61.92 & 27.42 & 28.45 \\
RAS & 5 & 25.0\% & 1.94 & \textbf{51.92} & \textbf{25.67} & \textbf{29.06} \\
RAS & 5 & 12.5\% & \textbf{1.99} & 53.24 & 26.04 & 28.94 \\

\hline
\textbf{Lumina} & & & & & & \\
\hline
\rowcolor{gray!10}
RFlow & 7 & 100\% & 0.49 & 48.19 & 38.60 & 28.65 \\
RAS & 10 & 25.0\% & 0.59 & \textbf{45.67} & \textbf{32.36} & \textbf{29.82} \\
RAS & 10 & 12.5\% & \textbf{0.65} & 47.34 & 32.69 & 29.75 \\
\hline
\rowcolor{gray!10}
RFlow & 5 & 100.\% & 0.69 & 96.53 & 59.26 & 26.03 \\
RAS & 7 & 25.0\% & 0.70 & \textbf{53.93} & \textbf{39.80} & \textbf{28.85} \\
RAS & 7 & 12.5\% & 0.74 & 54.62 & 40.23 & 28.83 \\
RAS & 6 & 25.0\% & 0.75 & 67.16 & 46.46 & 27.85 \\
RAS & 6 & 12.5\% & \textbf{0.78} & 67.88 & 45.88 & 27.83 \\
\hline


\hline
\end{tabular}
\caption{Pareto Improvements of rectified flow with \ourmethod{} on COCO Val2014 1024$\times$1024. Full experiment results are available in the Supplementary Material.}
\label{tab:comparison}
\end{table}

\subsection{Generation Benchmarks}

We conduct a comparative evaluation between \ourmethod{} and rectified flow baselines, which uniformly reduce the number of timesteps for all tokens during inference.  
To comprehensively assess performance, we experiment with multiple configurations of inference timesteps and selective sampling ratios.  
The results can be interpreted from two complementary perspectives.

\noindent \textbf{Pushing the Efficiency Frontier.}  
From the first perspective, \ourmethod{} extends the efficiency frontier by further reducing inference cost at each fixed number of timesteps offered by rectified flow.  
As shown in Figure~\ref{fig:teaser}(c)(d), we generate 10,000 images using dense inference across timestep settings ranging from 3 to 30, and subsequently apply \ourmethod{} with varying average sampling ratios over the selective-sampling stages.  
The results demonstrate that \ourmethod{} substantially decreases inference time with only marginal impact on generation quality metrics.  
For example, applying \ourmethod{} with a 25\% sampling ratio over 30 timesteps achieves a \textbf{2.25$\times$} increase in throughput, with only a \textbf{22.12\%} rise in FID, a \textbf{26.22\%} rise in sFID, and a negligible \textbf{0.065\%} drop in CLIP score.  
Moreover, the efficiency gains achieved through \ourmethod{} are obtained at a lower quality cost than those incurred by simply reducing the total number of sampling steps. Specifically, the rate of quality degradation when decreasing \ourmethod{}’s sampling ratio is significantly smaller than that observed when performing dense inference with fewer timesteps, particularly in low-step regimes ($<10$ steps).  
These findings highlight \ourmethod{} as a practical and effective strategy for improving inference efficiency while maintaining visual fidelity and prompt alignment.

\noindent \textbf{Pareto Improvements over Uniform Sampling.}  
We observe that \ourmethod{} frequently yields \emph{Pareto improvements} over rectified-flow baselines.  
To demonstrate this, we sorted the results from Stable Diffusion~3 and Lumina-Next-T2I by throughput and compared different configurations of \ourmethod{} with their closest baseline counterparts in Table~\ref{tab:comparison}.  
Across nearly all cases, \ourmethod{} achieves higher throughput while simultaneously improving FID, sFID, and CLIP scores compared with dense rectified-flow inference.  
These results indicate that for any given throughput level, \ourmethod{} provides configurations that offer both superior speed and quality, effectively expanding the Pareto frontier for balancing efficiency, fidelity, and text–image alignment.
\begin{table}
\fontsize{9pt}{10pt}\selectfont
\setlength\tabcolsep{4pt}
\centering
\begin{tabular}{lccc}
\multicolumn{4}{c}{\textbf{Stable Diffusion 3}} \\ \hline
\textbf{Method} & \textbf{Steps} & \textbf{Memory (GB)} & \textbf{Speedup} \\ \hline
RFlow & 28 & 19.21 (1x) & 1x \\
\rowcolor{gray!10} RAS-50\% & 28 & 20.36 (1.06x) & 1.62x \\
\rowcolor{gray!10} RAS-12.5\% & 28 & 20.36 (1.06x) & 2.44x \\ \hline
\end{tabular}

\vspace{0.5em}

\begin{tabular}{lccc}
\multicolumn{4}{c}{\textbf{Lumina-Next-T2I}} \\ \hline
\textbf{Method} & \textbf{Steps} & \textbf{Memory (GB)} & \textbf{Speedup} \\ \hline
RFlow & 30 & 10.30 (1x) & 1x \\
\rowcolor{gray!10} RAS-50\% & 30 & 10.73 (1.04x) & 1.56x \\
\rowcolor{gray!10} RAS-12.5\% & 30 & 10.73 (1.04x) & 2.70x \\ \hline
\end{tabular}

\caption{Memory Consumption of RAS.}
\label{tab:memory}
\end{table}
\subsection{Memory Consumption}

Since \ourmethod{} introduces caching of intermediate noise estimates as well as attention keys and values during inference, we quantify its additional memory overhead in Table~\ref{tab:memory}.  
\ourmethod{} incurs only a modest memory increase, approximately 6\% for Stable Diffusion~3 and 4\% for Lumina-Next-T2I, relative to baseline inference.  
This overhead is acceptable given the substantial acceleration gains achieved.  
Moreover, the additional memory consumption remains stable across different sampling ratios, as the full set of activations is cached for reuse throughout the inference process.
\begin{figure}[t]
    \centering
        \includegraphics[width=0.9\linewidth]{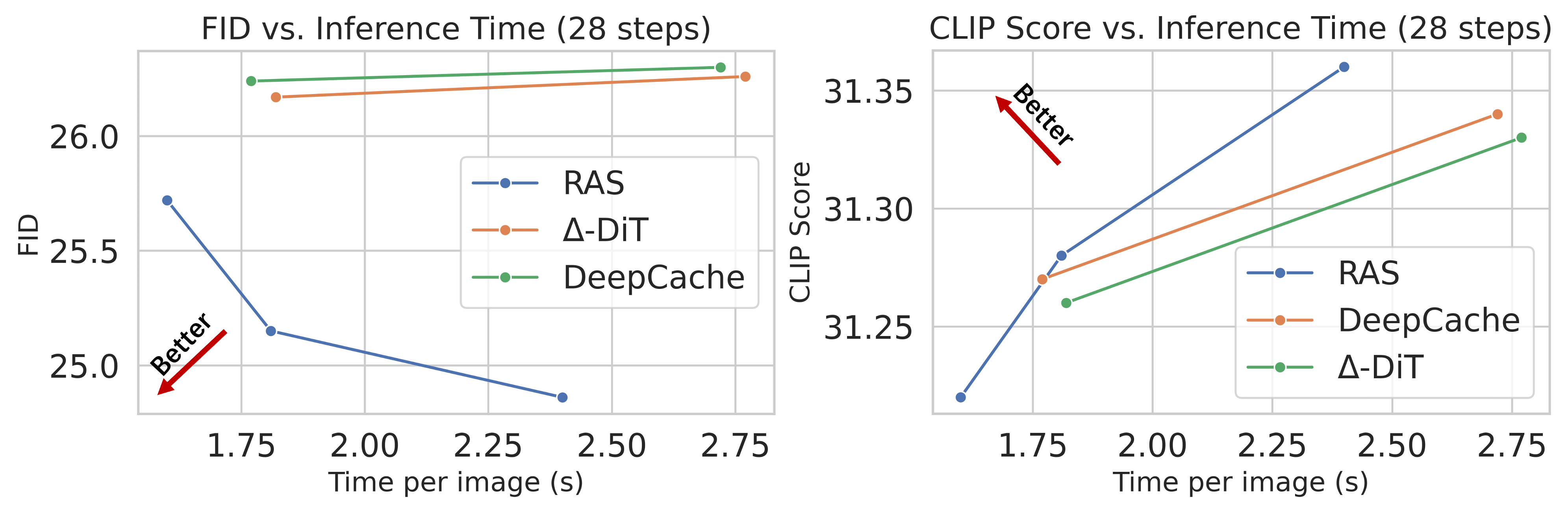}
    \vspace{-0.9em}
    \caption{Comparison on Stable Diffusion 3 of RAS with DeepCache and \(\Delta\)-DiT, utilizing different cache interval.}
    \label{fig:Comparison_RAS_DeepCache_DiT}
\end{figure}
\subsection{Comparison with Layer-Wise Methods}

Although orthogonal in principle, we compare \ourmethod{} against representative layer-wise cache-based acceleration methods for a more comprehensive analysis, using a subset of 5,000 images from MS-COCO.  
We manually adapt DeepCache~\cite{xu2018deepcache} for DiT models by reusing intermediate features and reimplement $\Delta$-DiT~\cite{chen2024delta} according to its official description.  
As shown in Figure~\ref{fig:Comparison_RAS_DeepCache_DiT}, \ourmethod{} achieves greater speedups while also yielding improved FID and CLIP scores.  
These results further highlight the advantage of region-adaptive token selection over layer-wise caching strategies in fully Transformer-based diffusion models.
\subsection{Detailed Prompts, Object Positions, and Counts}

To evaluate the robustness of \ourmethod{} under highly detailed prompts, particularly when users specify precise object counts, spatial arrangements, or relationships, we conduct experiments on the ParaImage-3000~\cite{wu2023paragraphtoimagegenerationinformationenricheddiffusion} and GenEval~\cite{ghosh2023genevalobjectfocusedframeworkevaluating} datasets.  
In the evaluations, \ourmethod{} maintains comparable overall performance to baseline inference and achieves Pareto improvements across multiple evaluation dimensions, confirming that region-adaptive sampling preserves compositional and positional accuracy even under complex generation constraints. Due to the space limit, please refer to the quantitative result tables in the supplementary material.
\subsection{Human Evaluation}

To further assess perceptual quality, we conduct a human evaluation to determine whether \ourmethod{} improves efficiency without compromising visual fidelity.  
We select 14 prompts from the official papers and blogs of Stable Diffusion~3 and Lumina-Next-T2I, generating two images per prompt, one using dense inference and the other using \ourmethod{} with the same random seed and timestep schedule.  
During the selective sampling phase, \ourmethod{} employs a 50\% average sampling ratio.  
We recruit 100 participants from 18 universities and companies to perform pairwise comparisons between outputs.  
Participants consistently report that images generated with \ourmethod{} are visually indistinguishable from or slightly preferred over those from dense inference, reinforcing the effectiveness of our approach in preserving perceptual quality while significantly improving inference throughput.

As shown in Figure~\ref{fig:teaser}(e), 45.21\% of 1,400 votes judged the two images to be of comparable quality, while 28.29\% favored the dense inference results and 26.50\% preferred outputs generated by \ourmethod{}.  
These results indicate that \ourmethod{} achieves substantial throughput gains: \textbf{1.625$\times$} on Stable Diffusion~3 and \textbf{1.561$\times$} on Lumina-Next-T2I, with negligible degradation in perceived image quality.

Furthermore, we evaluate \ourmethod{} on three human-preference–aligned metrics: ImageReward~\cite{xu2023imagerewardlearningevaluatinghuman}, PickScore~\cite{kirstain2023pickapicopendatasetuser}, and HPSv2~\cite{wu2023humanpreferencescorev2}.  As shown in Table~\ref{tab:human_preference}, \ourmethod{} maintains high performance across all benchmarks while delivering significantly faster generation, reaffirming that our region-adaptive sampling achieves human-aligned quality at markedly lower computational cost.

\subsection{Ablation Study}
\begin{table}

\fontsize{9pt}{10pt}\selectfont
\setlength\tabcolsep{2pt}
\centering

\begin{tabular}{lccc}
    \multicolumn{4}{c}{(a) Drop Scheduling} \\
    \hline
    \textbf{Method} & \textbf{FID $\downarrow$} & \textbf{sFID $\downarrow$} & \textbf{CLIP score $\uparrow$} \\
    \hline
    \rowcolor{gray!10}
    Default & 35.81 & 18.41 & 30.13 \\
    Static Sampling Freq. & 37.92 & 19.11 & 29.98 \\
    Random Dropping & 43.19 & 22.23 & 29.65 \\
    W/O Error Reset & 46.10 & 24.85 & 30.41 \\
    \hline
\end{tabular}

\vspace{0.2em}

\begin{tabular}{lcccc}
    \multicolumn{5}{c}{(b) Key and Value Caching} \\
    \hline
    \textbf{Method} & \textbf{Timesteps} & \textbf{FID $\downarrow$} & \textbf{sFID $\downarrow$} & \textbf{CLIP score $\uparrow$} \\
    \hline
    \rowcolor{gray!10}
    Default & 28 & 24.30 & 26.26 & 31.34 \\
    W/O & 28 & 31.36 & 20.19 & 31.29 \\
    \hline
    \rowcolor{gray!10}
    Default & 10 & 35.81 & 18.41 & 30.13 \\
    W/O & 10 & 32.33 & 20.21 & 30.27 \\
    \hline
\end{tabular}

\vspace{0.2em}

\setlength\tabcolsep{1pt}

\hspace{-1mm}

\begin{minipage}[t]{0.46\linewidth}
\centering
\begin{tabular}{cccc}
    \multicolumn{4}{c}{(c) Error Reset Schedule} \\
    \hline
    \scalebox{0.8}{\textbf{Reset ID}} & 
    \scalebox{0.8}{\textbf{FID $\downarrow$}} & 
    \scalebox{0.8}{\textbf{sFID $\downarrow$}} & 
    \scalebox{0.8}{\textbf{CLIP $\uparrow$}} \\
    \hline
    5 & 27.04 & 19.03 & 31.33 \\
    \rowcolor{gray!10}
    8 & 24.60 & 17.24 & 31.31 \\
    11 & 25.80 & 16.67 & 31.17 \\
    7,11 & 24.58 & 15.82 & 31.31 \\
    \hline
\end{tabular}
\end{minipage}
\hspace{-1mm}
\begin{minipage}[t]{0.46\linewidth}
\centering
\begin{tabular}{lcccc}
    \multicolumn{5}{c}{(d) Starvation Prevention} \\
    \hline
    \scalebox{0.8}{\textbf{Method}} & 
    \scalebox{0.8}{\textbf{Steps}} & 
    \scalebox{0.8}{\textbf{FID} $\downarrow$} & 
    \scalebox{0.8}{\textbf{sFID} $\downarrow$} & 
    \scalebox{0.8}{\textbf{CLIP} $\uparrow$} \\
    \hline
    \rowcolor{gray!10}
    Default & 10 & 35.81 & 18.41 & 30.13 \\
    W/O & 10 & 39.87 & 19.75 & 29.84 \\
    \hline
    \rowcolor{gray!10}
    Default & 14 & 26.48 & 18.14 & 31.18 \\
    W/O & 14 & 26.58 & 17.96 & 31.11 \\
    \hline
\end{tabular}
\end{minipage}

\caption{Ablation Study on Stable Diffusion 3. All techniques including dynamic sampling ratio, region identifying, error reset, key \& value recovery are necessary for high quality generation.}
\label{tab:ablation_combined}
\end{table}
\noindent \textbf{Token Drop Scheduling.}  
As reported in Table~\ref{tab:ablation_combined}(a), we evaluate the impact of the scheduling strategies introduced in Section~\ref{sec.method}, including (1) dynamic sampling-ratio scheduling, (2) selective caching of dropped tokens, and (3) insertion of dense steps to reset accumulated errors.  
All experiments are conducted on Stable Diffusion~3 using 10 timesteps and an average sampling ratio of 12.5\%.  
Results show that each component contributes positively to image quality, with the combination providing the best overall balance between efficiency and fidelity.

\noindent \textbf{Key–Value Caching.}  
As shown in Table~\ref{tab:ablation_combined}(b), reusing keys and values from the previous timestep is critical for maintaining generation quality, particularly when using longer sampling schedules.  
While discarding the keys and values of inactive tokens slightly improves throughput, it substantially distorts the attention distribution of active tokens.  
Importantly, a low ranking in our region metric does not imply negligible influence in the attention mechanism—demonstrating the necessity of caching to preserve cross-token contextual integrity.

\noindent \textbf{Error-Resetting Schedule.}  
Table~\ref{tab:ablation_combined}(c) presents results on different error-resetting schedules using 14 timesteps on Stable Diffusion~3.  
We find that inserting a dense-step reset near the midpoint of the selective-sampling phase (e.g., between steps 4–13) yields the optimal trade-off between image quality and computation.  
Adding additional dense steps offers only marginal improvement while introducing nontrivial time overhead.

\noindent \textbf{Starvation Prevention.}  
Finally, Table~\ref{tab:ablation_combined}(d) validates the effectiveness of our starvation-prevention mechanism.  
By tracking token drop frequency and scaling their reactivation probability, this component prevents persistent omission of background regions and stabilizes long-horizon sampling—achieving quality improvements with negligible computational overhead.
\begin{table}

\centering
\fontsize{9pt}{10pt}\selectfont
\setlength\tabcolsep{2pt}
\begin{tabular}{ccccccc}
    \hline
    \scalebox{0.8}{\textbf{Method}} & 
    \scalebox{0.8}{\textbf{Steps}} & 
    \scalebox{0.8}{\textbf{Time(s)}} & 
    \scalebox{0.8}{\textbf{SpeedUp $\uparrow$}} &
    \scalebox{0.8}{\textbf{Img. Rew. $\uparrow$}} &
    \scalebox{0.8}{\textbf{PickScore $\uparrow$}} & 
    \scalebox{0.8}{\textbf{hpsv2 $\uparrow$}} \\
    \hline
    RFlow & 30 & 8.77 & 1 & 0.37 & 21.88 & 0.26 \\
    \hline
    RFlow & 15 & 4.36 & 2.01 & 0.13 & 21.45 & 0.24 \\
    \cellcolor{gray!10}RAS-25\% & \cellcolor{gray!10}30 & \cellcolor{gray!10}3.89 & \cellcolor{gray!10}2.26 & \cellcolor{gray!10}\textbf{0.13} & \cellcolor{gray!10}\textbf{21.45} & \cellcolor{gray!10}0.22 \\
    \cellcolor{gray!10}RAS-75\% & \cellcolor{gray!10}15 & \cellcolor{gray!10}3.72 & \cellcolor{gray!10}2.35 & \cellcolor{gray!10}0.05 & \cellcolor{gray!10}21.34 & \cellcolor{gray!10}\textbf{0.24} \\
    \hline
    RFlow & 10 & 2.92 & 3 & -0.20 & 20.94 & 0.21 \\
    \cellcolor{gray!10}RAS-25\% & \cellcolor{gray!10}15 & \cellcolor{gray!10}2.31 & \cellcolor{gray!10}3.78 & \cellcolor{gray!10}\textbf{-0.18} & \cellcolor{gray!10}\textbf{20.98} & \cellcolor{gray!10}\textbf{0.21} \\
    \hline
    RFlow & 7 & 2.05 & 4.27 & -0.75 & 20.24 & 0.19 \\
    \cellcolor{gray!10}RAS-25\% & \cellcolor{gray!10}10 & \cellcolor{gray!10}1.70 & \cellcolor{gray!10}5.15 & \cellcolor{gray!10}\textbf{-0.43} & \cellcolor{gray!10}\textbf{20.54} & \cellcolor{gray!10}\textbf{0.19} \\
    \cellcolor{gray!10}RAS-12.5\% & \cellcolor{gray!10}10 & \cellcolor{gray!10}1.54 & \cellcolor{gray!10}5.68 & \cellcolor{gray!10}\textbf{-0.54} & \cellcolor{gray!10}\textbf{20.34} & \cellcolor{gray!10}0.18 \\
    
    \hline
\end{tabular}
\vspace{-1em}

\caption{Benchmarks for evaluating the human preference on Lumina-Next-T2X. RAS-X\% stands for RAS with X\% tokens activated each step. RAS provides Pareto improvements in multiple settings.}
\label{tab:human_preference}
\end{table}
\section{Conclusion}
\label{sec.conclusion}

In this work, we investigate the spatial heterogeneity in diffusion generation, observing that different image regions demand varying refinement levels during denoising and exhibit strong temporal continuity across steps.  
Leveraging these insights, we propose \textbf{\ourmethod{}}, a training-free sampling strategy that dynamically allocates computation based on regional attention—prioritizing semantically important areas while reusing cached predictions for less critical ones.  

Extensive experiments and user studies show that \ourmethod{} achieves significant acceleration with minimal perceptual loss, consistently surpassing uniform sampling baselines.  
These results underscore the promise of region-adaptive and temporally aware strategies for efficient diffusion transformers and real-time generative modeling.

{
    \small
    \bibliographystyle{ieeenat_fullname}
    \bibliography{main}
}


\end{document}